\documentclass[letterpaper, 10pt, conference, twoside]{IEEEtran}
\usepackage[letterpaper, top=54pt, bottom=54pt, left=54pt, right=54pt]{geometry}
\def\IEEEtitletopspace{18pt}

\IEEEoverridecommandlockouts
\usepackage{cite}
\usepackage{amsmath,amssymb,amsfonts}
\usepackage{algorithmic}
\usepackage{graphicx}
\usepackage{textcomp}
\usepackage{xcolor}
\def\BibTeX{{\rm B\kern-.05em{\sc i\kern-.025em b}\kern-.08em
    T\kern-.1667em\lower.7ex\hbox{E}\kern-.125emX}}

\usepackage{xspace}
\newcommand{\eg}{\emph{e.g., }}
\newcommand{\etal}{\emph{et al.}\xspace}

\usepackage{lipsum}
\usepackage[nolist]{acronym}
\begin{acronym}
\acro{cnn}[CNN]{Convolutional Neural Network}
\acro{mot}[MOT]{Multi-Object Tracking}
\acro{mpt}[MPT]{Multi-Pedestrian Tracking}
\acro{mtt}[MTT]{Multi-Target Tracking}
\acro{jde}[JDE]{Joint Detection and Embedding}
\acro{sde}[SDE]{Separate Detection and Embedding}
\acro{iou}[IoU]{Intersection over Union}
\acro{tbd}[TBD]{Tracking-by-Detection}
\acro{reid}[ReID]{Re-Identification}
\acro{roi}[RoI]{Region of Interest}
\acro{giou}[GIoU]{Generalized Intersection over Union}
\acro{fifo}[FIFO]{First-In, First-Out}
\acro{mgn}[MGN]{Multi-Granularity Network}
\acro{bn}[BN]{Batch Normalization}
\acro{hota}[HOTA]{Higher Order Tracking Accuracy}
\acro{mota}[MOTA]{Multiple Object Tracking Accuracy}
\acro{vit}[ViT]{Vision Transformer}
\acro{dl}[DL]{Deep Learning}
\acro{ai}[AI]{Artificial Intelligence}
\acro{amr}[AMR]{Autonomous Mobile Robot}
\acro{map}[mAP]{mean Average Precision}
\acro{fps}[FPS]{Frames Per Second}
\acro{ema}[EMA]{Exponential Moving Average}
\end{acronym}
\usepackage{color,soul}
\usepackage{booktabs}
\usepackage{balance}
\usepackage{soul}
\usepackage{censor}
\usepackage{fancyhdr}
\usepackage{hyperref}

\begin{document}

% --- Headers ---
\pagestyle{fancy}
\fancyhf{}

\fancyhead[LE,RO]{\footnotesize \thepage}
\fancyhead[LO,RE]{\footnotesize IEEE RO-MAN 2026. PREPRINT VERSION. ACCEPTED MAY 2026.}

\renewcommand{\headrulewidth}{0pt}
% --- Headers ---

\title{\LARGE \bf Does Appearance Help? A Systematic Study of Image-Based Re-Identification in Online 3D Multi-Pedestrian Tracking\\
}

\author{Eduardo Borges, Luís Garrote, and Urbano J. Nunes% <-this % stops a space
    \thanks{Authors are with the Institute of Systems and Robotics, Department of Electrical and Computer Engineering, University of Coimbra, Portugal.
            {\tt\small \{eduardo.borges, garrote, urbano\}@isr.uc.pt}}%
}

\maketitle
\thispagestyle{fancy}

\begin{abstract}
LiDAR-based 3D Multi-Object Tracking (MOT) typically relies solely on geometric information, which is often insufficient to distinguish between targets during prolonged occlusions or in crowded human-populated environments. While integrating RGB-based Re-Identification (ReID) offers a theoretical solution for preserving identity context, existing approaches often rely on computationally expensive parallel detectors that hinder real-time robot responsiveness. This work presents a systematic study of image-based ReID in online 3D MOT, utilizing a lightweight projection-based framework to decouple geometric and appearance modeling for mobile robots. A comprehensive analysis of feature extraction architectures is conducted, employing lightweight CNNs and Vision Transformers, and evaluating various multi-modal data association strategies to balance computational latency with robust tracking. Experiments on the \textit{Pedestrian} class of the KITTI dataset reveal that naive linear fusion, of appearance and motion costs, degrades performance due to visual noise. Conversely, a cascaded matching strategy successfully recovers occluded tracks without compromising overall precision, effectively preventing identity switches to maintain human-robot interaction continuity. We show that lightweight architectures can offer an optimal trade-off between the low latency required for safe navigation and the discriminative power needed for social awareness. 

\end{abstract}


\section{Introduction}

The deployment of social and service robots has been a rapidly growing field, particularly in the last decade, mainly due to fast and impactful advancements in deep learning and artificial intelligence~\cite{grigorescu2020survey}. These advancements enabled the development of models and techniques that are increasingly effective in accomplishing tasks related to perception, localization, and human behavior analysis. These tasks are essential to ensure the safe and socially acceptable navigation of an agent (\eg a service robot or an autonomous mobile robot) within dynamic, human-populated environments such as crowded public spaces, hospitals, or factory floors.

To facilitate natural human-robot interaction and socially aware navigation, the agent must not only detect surrounding pedestrians but also track their movement over time. This allows the robot to predict human trajectories, respect personal space, and maintain interaction context. \ac{mot} addresses this by first generating detections from sensor data. Second, it assigns a unique identifier (ID) to each person, associating detections across frames to manage stored trajectories. In the context of human-robot interaction, reliably maintaining this unique ID is critical. This ensures the agent does not lose track of an interaction partner during prolonged occlusions in crowded environments.

LiDAR has become one of the most widely used sensors for 3D \ac{mot}, mainly due to its precise depth measurements~\cite{shi2019pointrcnn}. This motivated many trackers, such as AB3DMOT~\cite{weng2020ab3dmot}, to rely solely on geometric information (\eg 3D \ac{giou}~\cite{rezatofighi2019generalized}) and motion prediction (\eg Kalman Filter~\cite{kalman1960anew}) for data association. These choices lead to highly computationally efficient systems. However, motion-only approaches face challenges in complex scenarios. In crowded scenes, where pedestrians are close and occlusions are frequent or prolonged, geometric information is often insufficient to distinguish between pedestrians or to identify lost targets, leading to identity switches or fragmented tracks. Here, knowledge of the object's appearance is required.
\begin{figure}[t]
   \centering
   \includegraphics[width=1\linewidth]{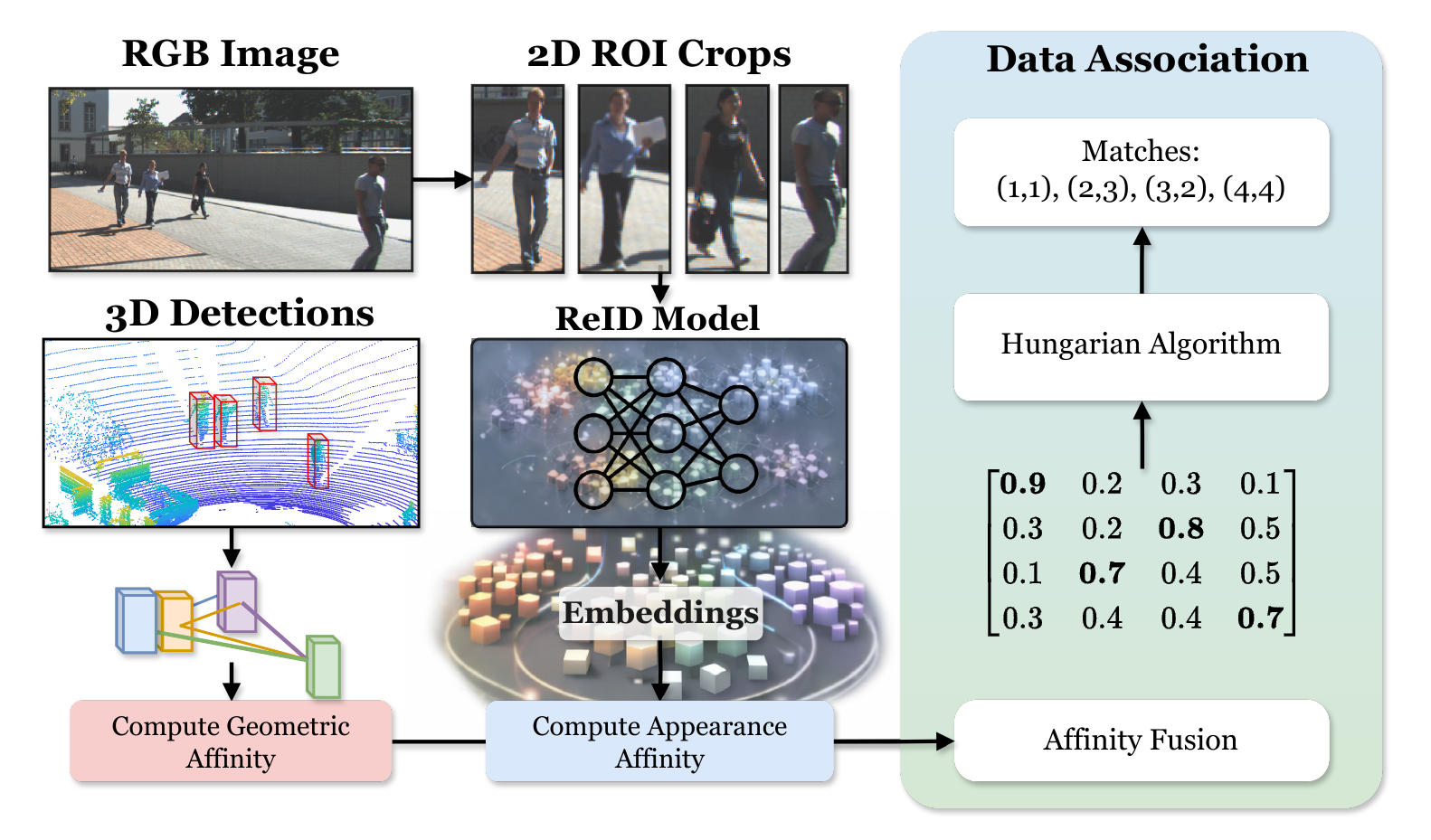}
   \caption{\textbf{Overview of the proposed re-identification framework.} The pipeline integrates an additional RGB-based appearance branch to complement motion with appearance information in data association.}
   \label{fig:systemoverview}
\end{figure}

In this work, a comprehensive study was conducted to analyze the impact of image-based \ac{reid} on online 3D \ac{mpt}. A lightweight baseline was established with PointPillars~\cite{lang2019pointpillars} as the 3D detector and AB3DMOT~\cite{weng2020ab3dmot} as the tracker. Then, a modular projection-based \ac{reid} branch was integrated to introduce appearance information. To select an appropriate \ac{reid} network, a range of feature extractors was evaluated, from lightweight \acp{cnn} to \acp{vit}. The overall pipeline is illustrated in Fig.~\ref{fig:systemoverview}. 

This work's contributions are summarized as follows: 
\begin{itemize}
    \item Systematic ReID Benchmark: A comprehensive set of feature-extracting backbones is benchmarked to determine the optimal trade-off between inference latency and discriminative power for online 3D \ac{mot}. 
    \item Robust Fusion Strategy: Different multi-modal association strategies are evaluated to identify the optimal balance between motion and appearance.
    \item Appearance Modeling Strategy: Different appearance modeling strategies are evaluated to identify the most effective method to preserve appearance information.
    \item Impact Analysis: An analysis on the KITTI MOT dataset is provided using common metrics such as HOTA, MOTA, and IDF1. Finally, the impact of domain adaptation and appearance memory strategies on tracking robustness is analyzed. 
\end{itemize}

\section{Related Work}

\subsection{Multi-Object Tracking}
Within \ac{mot}, one of the most popular paradigms follows the \ac{tbd} approach, where the process is separated into two independent stages of object detection and temporal association~\cite{hassan2024multi, guan2025multi}. Some adopters of this paradigm, such as SORT~\cite{bewley2016sort} and AB3DMOT~\cite{weng2020ab3dmot}, for 2D and 3D \ac{mot}, respectively, focused on maximizing speed needed for online deployment, relying solely on motion information. Both utilize \ac{iou} to calculate the spatial overlap between current detections and Kalman Filter predictions based on previous detections. Then, they use the Hungarian algorithm~\cite{kuhn1955hungarian} to perform the optimal assignment between them. Although effective for short-term tracking on non-challenging datasets, motion-only systems~\cite{bewley2016sort, weng2020ab3dmot,zhang2022bytetrack,cao2023ocsort} suffer in crowded scenes or during prolonged occlusions. Wojke \etal extended the original SORT framework by introducing DeepSORT~\cite{wojke2017deepsort}, a variant that integrates a lightweight \ac{cnn} to extract appearance-based features. This addition enabled object re-identification following temporary disappearance or occlusion, marking a shift in MOT systems toward a greater emphasis on appearance-based association. 

\begin{figure}[t]
    \centering
    \includegraphics[width=1\linewidth]{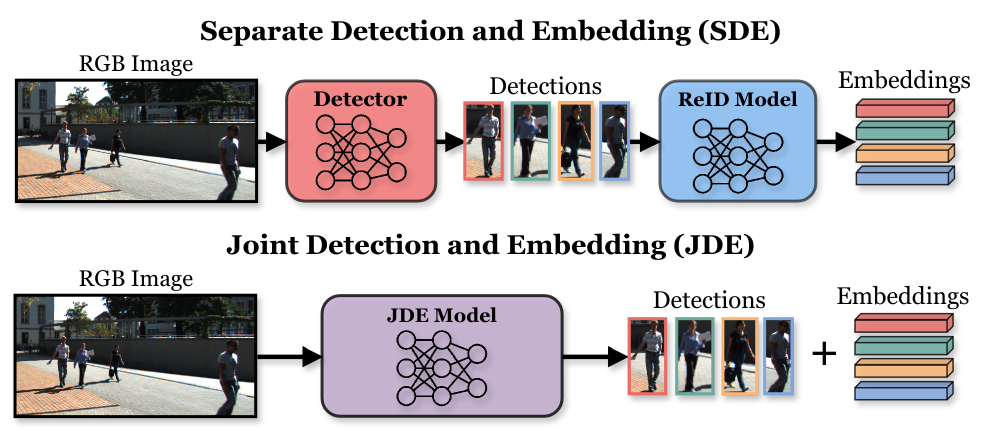}
    \caption{\textbf{Separate Detection and Embedding (SDE) and Joint Detection and Embedding (JDE) paradigms.} \ac{sde} employs two independent networks, maximizing accuracy. \ac{jde} integrates both tasks into a single backbone, achieving higher speed but requiring a careful balancing of the loss functions.}
    \label{fig:sdejde}
\end{figure}

This paradigm shift gave rise to two new approaches, highlighted in Fig.~\ref{fig:sdejde}, focused on the first stage of \ac{mot}: \ac{sde}, in which object detection and embedding extraction for \ac{reid} are handled by independent networks, and \ac{jde}, where both tasks are performed by a single network. \ac{sde} frameworks~\cite{wojke2017deepsort, shenoi2020jrmot, zhang2021drop, du2023strongsort, maggiolino2023deepocsort, lee2024dinomot} benefit from the ability to utilize high-performance specialized backbones. However, because \ac{reid} must be performed every frame, these often suffer from computational redundancy and lower speeds. \ac{jde} frameworks~\cite{chan2022jde, zhang2021fairmot, liang2022adamot, liang2022cstrack, li2022simpletrack, yang2022dpanet} achieve higher speeds due to the integration of both tasks into a single network. Nevertheless, they typically exhibit inferior performance and must carefully address task competition between class-level recognition for detection and instance-level discrimination for \ac{reid}.

\subsection{Object Re-Identification}
A standard \ac{reid} pipeline typically involves designing or selecting an appropriate loss function to train a \ac{cnn}-based feature extraction backbone. Some methods employ well-established networks such as ResNet~\cite{ResNet} or MobileNet~\cite{MobileNet}, while others propose task-specific \ac{cnn} architectures tailored for \ac{reid}. For example, MGN~\cite{wang2018mgn} modifies a \ac{cnn} backbone to simultaneously learn global representations and fine-grained part-level features, improving robustness to pose and viewpoint variations. More recently, \acp{vit} have emerged as a capable alternative. Unlike \acp{cnn}, which rely on local information, transformers leverage self-attention mechanisms to capture long-range dependencies and global context, offering increased robustness in handling occlusions and complex pose variations~\cite{he2021transreid}.

Regardless of the architecture, some of the most commonly used loss functions to train \ac{reid} networks are the cross-entropy loss, the triplet loss~\cite{Triplet_Loss}, and the contrastive loss. Cross-entropy loss treats \ac{reid} as a classification problem by training the network to assign a class to each object or identity. The learned feature space implicitly clusters the embeddings of the same object. Both contrastive and triplet losses follow the same principle: minimize the distance between embeddings of the same object (positive pairs), while simultaneously maximizing the distance between embeddings of different objects (negative pairs). However, they differ in their implementation: contrastive loss operates on \textit{pairs} of samples, while triplet loss works with \textit{triplets} of samples (anchor, positive, and negative). Recent advancements, such as proposed by Luo \etal~\cite{luo2019bnneck}, have further refined these pipelines by introducing techniques such as the BNNeck and warm-up learning rate schedules, establishing strong baselines that balance the optimization of both cross-entropy and triplet losses.

\begin{figure*}[t]
    \centering
    \includegraphics[width=1\linewidth]{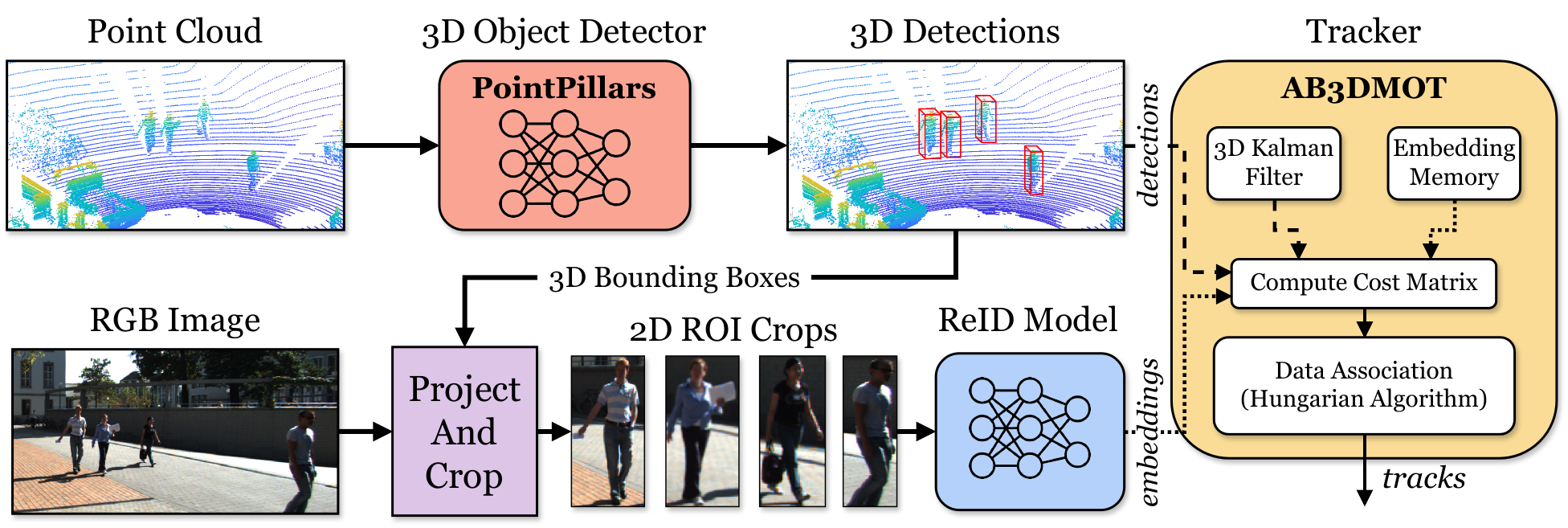}
    \caption{\textbf{Overview of the proposed Multi-Modal 3D MOT framework.} The pipeline operates in two parallel streams. The geometric branch (top) processes the raw LiDAR point cloud using a \textbf{PointPillars} detector to generate 3D bounding boxes. The appearance branch (bottom) projects these 3D detections onto the RGB image to extract the corresponding 2D \ac{roi} crops, which are then encoded by a \ac{reid} network (\textit{embeddings}). Finally, a modified \textbf{AB3DMOT} tracker fuses these modalities, utilizing the Hungarian Algorithm to perform the data association, outputting the final updated trajectories (\textit{tracks}).}
    \label{fig:system_overview}
\end{figure*}

\subsection{Multi-Modal 3D Tracking}
To overcome the sparsity of LiDAR data, recent approaches have proposed the integration of information from RGB images. A common approach, used by state-of-the-art frameworks like EagerMOT~\cite{kim2021eagermot} and DeepFusionMOT~\cite{wang2022deepfusionmot}, relies on a ``decision-level fusion'' strategy. These systems utilize two independent detectors running in parallel: a 3D detector for the point cloud and a 2D detector for the RGB image. The 2D and 3D detections are then fused, allowing the 3D tracker to inherit the robustness of 2D trackers (usually using appearance embeddings). While effective, this dependency on two detectors increases the computational complexity. In contrast, fewer works~\cite{lee2024dinomot} explore a projection-based \ac{sde} approach, where 3D detections are directly projected to extract image features, which avoids a dedicated 2D detection pipeline while still capturing appearance information.

\subsection{Impact of Re-Identification on MOT Performance}
Despite the prevalence of appearance models in modern trackers, research quantifying their specific contribution remains sparse. Early metrics like \ac{mota}~\cite{bernardin2008evaluating} are heavily biased toward detection accuracy and often fail to capture improvements in association. In contrast, identity-aware metrics such as IDF1~\cite{ristani2016performance} and the more recently proposed \ac{hota}~\cite{luiten2021hota} provide a clearer measure of tracking consistency, penalizing identity switches explicitly. Several ablation studies~\cite{zhang2022bytetrack, cao2023ocsort} have claimed that in high frame rate scenarios, simple motion heuristics (like IoU or Kalman Filters) can outperform complex \ac{reid} models, which may introduce noise due to appearance ambiguities. Conversely, in scenarios with prolonged occlusions or low frame rates, \ac{reid} has been shown to be critical for track recovery~\cite{wojke2017deepsort, zhang2022bytetrack}. This highlights the need for careful ablation studies, such as the one presented in this work, to clearly differentiate between the gains of geometric association and those of visual discrimination.

\section{Methodology}
\subsection{System Overview}
The proposed 3D \ac{mot} framework follows a \ac{tbd} paradigm, integrating motion information from LiDAR point clouds with appearance features from RGB images to enhance tracking robustness. The overall architecture of the framework is illustrated in Fig.~\ref{fig:system_overview}.

The system operates in four main stages. First, a 3D object detector (PointPillars~\cite{lang2019pointpillars}) processes the raw LiDAR point cloud to generate 3D bounding box proposals. Second, these 3D detections are projected onto the corresponding RGB image plane using the camera calibration matrices. This projection allows for the extraction of 2D \ac{roi} crops corresponding to each detected 3D object. Third, the extracted image crops are fed into a \ac{reid} network. This module maps each crop into a high-dimensional embedding space, where the distance between embeddings represents the appearance similarity between pedestrians. Finally, the tracking module, built on top of the AB3DMOT~\cite{weng2020ab3dmot} framework, performs state estimation and data association. The association step combines motion information (via 3D \ac{giou}~\cite{rezatofighi2019generalized}) with the extracted appearance embeddings (via cosine distance) to match current detections with existing tracklets, to minimize identity switches in complex scenarios.

\subsection{RGB-Based ReID Module}

\subsubsection{Feature Extraction Architectures}
\begin{figure}[t]
    \centering
    \includegraphics[width=1\linewidth]{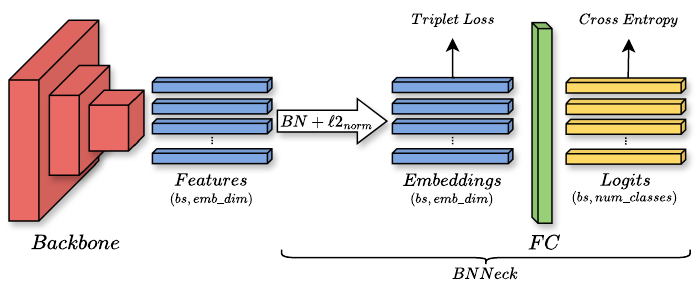}
    \caption{\textbf{Architecture of the ReID head employing the BNNeck\cite{luo2019bnneck} strategy.} Features extracted from the backbone are processed through a combined \acf{bn} and $\ell_2$ normalization block. The resulting normalized embeddings are utilized for both metric learning (Triplet Loss) and inference during tracking, while a final fully connected (FC) layer maps them to class logits for the Cross-Entropy loss.}
    \label{fig:bnneck}
\end{figure}

To identify the optimal trade-off between feature representativeness and inference speed for 3D tracking, a diverse set of backbone architectures was evaluated. These were categorized into four groups:
\begin{itemize}
    \item \textbf{Standard CNNs:} ResNet-18 and ResNet-50~\cite{ResNet} were selected as robust baselines for feature extraction.
    \item \textbf{Lightweight CNNs:} To address the computational constraints of real-time deployment, MobileNetV2 and MobileNetV3-Small~\cite{MobileNet} were evaluated.
    \item \textbf{Vision Transformers:} ViT-B/16, ViT-B/32~\cite{dosovitskiy2020vit}, and Swin-T~\cite{liu2021swin} were explored due to the growing popularity of feature extraction transformer-based architectures and to assess the impact of attention mechanisms.
    \item \textbf{Specialized ReID Architectures:} The MGN architecture~\cite{wang2018mgn} was implemented due to its fine-grained feature retrieval capability. To maintain system efficiency, the standard MGN backbone was replaced with the lightweight MobileNetV3-Small.
\end{itemize}
For all architectures, the original classification heads were substituted with a custom projection head, following the \textit{BNNeck} strategy proposed by Luo \etal~\cite{luo2019bnneck}. First, the backbone features are mapped to a lower-dimensional embedding space of size $d$ using a linear layer. This is followed by a one-dimensional \ac{bn} layer and an $\ell_2$ normalization operation:
\begin{equation}
    f_{norm} = \frac{\text{BN}(W_{proj} \cdot f_{backbone})}{\| \text{BN}(W_{proj} \cdot f_{backbone}) \|_2}
\end{equation}
Normalized embeddings $f_{norm}$ are used for the calculation of triplet loss and during inference. Finally, a fully connected classifier layer maps these normalized embeddings to class logits for the cross-entropy loss calculation. The complete \ac{reid} framework is illustrated in Fig.~\ref{fig:bnneck}.

\subsubsection{Training Objective}
The networks were optimized using a loss function designed to ensure both within-class compactness and between-class separability. The total loss $L_{total}$ is defined as the sum of the classification loss and the metric learning loss:
\begin{equation}
    L_{total} = L_{ce} + L_{triplet}
\end{equation}
where $L_{ce}$ is the Cross-Entropy loss computed on the class logits, and $L_{triplet}$ is the Triplet Margin loss with hard negative mining. 

\subsubsection{Embedding Dimensionality}
The size of the embedding vector $d$ is a hyperparameter that affects both discriminability and memory size. To determine the most efficient configuration, an ablation study was performed across five different embedding sizes: $d \in \{128, 256, 512, 1024, 2048\}$. This allows for the selection of a dimension that maximizes \ac{reid} accuracy while minimizing the computational cost of the data association step.

\subsubsection{Domain Adaptation}
A notable challenge when applying \ac{reid} to 3D MOT is the domain gap between standard surveillance viewpoints and the ego-perspective of a mobile robot sharing the floor with humans. To address this, a two-stage training strategy was employed:

\begin{enumerate}
    \item \textbf{General Pre-training:} Models were first trained on the large-scale Market-1501 dataset~\cite{Liang2015market1501}, allowing the network to learn robust general-purpose pedestrian features.
    \item \textbf{Target Domain Fine-tuning:} To adapt the representations to the specific viewpoints, lighting, and occlusions found in human-robot shared environments, some selected models were fine-tuned in an adapted version of the KITTI~\cite{Geiger2012kitti} dataset. This \textit{KITTI-ReID} dataset was constructed by extracting 2D pedestrian crops from RGB images, using the ground-truth track IDs as unique identities.
\end{enumerate}

\subsection{Tracking and Association}

\subsubsection{State Estimation}
A standard Kalman Filter framework was employed to estimate the state of each tracked pedestrian in 3D space, following the formulation of AB3DMOT~\cite{weng2020ab3dmot}. The state vector of a tracklet is defined as $s = (x, y, z, \theta, l, w, h, v_x, v_y, v_z)$, representing the 3D center coordinates, heading angle, bounding box dimensions, and linear velocities, respectively.
A constant velocity model is employed for the state transition, allowing the tracker to predict the pedestrian's position in the next frame. When a detection is associated with a tracklet, the Kalman Filter update step corrects the predicted state using the observed detection parameters. If no detection is matched for a predefined number of frames ($max\_age$), the tracklet is considered lost and removed from the system.

\subsubsection{Temporal Appearance Modeling}
An appearance model is essential for comparing and associating pedestrians over time, particularly to recover identities after occlusion. Three strategies to maintain and update the pedestrian-specific appearance representations were evaluated:

\paragraph{Fixed-Window Gallery} A fixed-size gallery $G = \{f_1, f_2, \dots, f_N\}$ is maintained for each tracklet to store the $N$ most recent embeddings. Embeddings are inserted in temporal order. Once the capacity $N$ is reached, a \ac{fifo} strategy is applied, removing the oldest feature to accommodate the new one.
During association, the similarity between a detection embedding $f_{det}$ and a tracklet is computed using a maximum similarity strategy:
\begin{equation}
    S_{gallery}(f_{det}, G) = \max_{f_i \in G} (\text{Sim}(f_{det}, f_i))
\end{equation}
This approach allows the tracker to match against multiple records of the pedestrian, increasing robustness to changes in viewing angle and pedestrian pose.

\paragraph{Cumulative Moving Average (CMA)} Instead of storing multiple vectors, a single template embedding $\mu_t$ is maintained representing the mean of all observed embeddings. The template is updated online as a cumulative moving average. For a tracklet with current mean $\mu_{t-1}$ and count $n$, the new mean after matching detection $f_{det}$ is:
\begin{equation}
    \mu_t = \frac{n-1}{n}\mu_{t-1} + \frac{1}{n}f_{det}
\end{equation}
This strategy provides a stable representation but may lag in adapting to rapid appearance changes.

\paragraph{Exponential Moving Average (EMA)} To improve stability, the pedestrian template $e_t$ is updated using an exponential moving average. Upon a match with detection $f_{det}$, the template is updated as:
\begin{equation}
    e_t = \alpha f_{det} + (1-\alpha)e_{t-1}
\end{equation}
where $\alpha$ is a momentum term. This gives greater weight to recent appearance features while retaining a decreasing memory of historical information.

\subsubsection{Multi-Modal Data Association}
To determine the optimal match between incoming 3D detections and existing tracklets, the tracking module must effectively fuse spatial proximity with visual appearance. We evaluate two distinct multi-modal fusion strategies for computing the final assignment cost:
\paragraph{Weighted Linear Combination} A fused cost matrix $C$ was constructed by linearly combining geometric and appearance costs. Geometric cost is defined as $C_{geo} = 1 - \text{GIoU}_{3D}$, and appearance cost as $C_{app} = 1 - S_{reid}$, where $S_{reid}$ is the cosine similarity score derived from the selected appearance model. The final cost matrix is as follows:
\begin{equation}
    C_{total} = w \cdot C_{geo} + (1-w) \cdot C_{app}
\end{equation}
The contribution of each cost can be adjusted by varying the weighting parameter $w$ within the range $[0,1]$, simultaneously considering spatial overlap and pedestrian appearance. The optimal assignment is obtained by applying the Hungarian algorithm to $C_{total}$. Assignments exceeding the cost threshold are discarded.

\paragraph{Cascaded Matching Strategy} Alternatively, the association is decomposed into a 2-step cascaded process. First, the assignment problem is solved using the geometric cost matrix $C_{geo}$ to associate spatially consistent detections.
Second, a new cost matrix $C_{app}$ is constructed from unmatched detections and unmatched tracklets obtained from the first stage. A second round of Hungarian matching is performed based purely on appearance similarity, allowing the system to recover tracks that have moved unexpectedly, effectively serving as an occlusion recovery mechanism.

\section{Experiments}

\subsection{Implementation Details}

\textbf{Experimental Setup:} Models were implemented using PyTorch and trained on a single NVIDIA RTX 5090 GPU. The \ac{reid} backbones were trained using the AdamW optimizer with a learning rate of $1 \cdot 10^{-3}$, a batch size of 32 and a triplet loss margin of 1, for 400 epochs. A 10-epoch warm-up strategy was applied followed by a plateau-based scheduler. For the 3D tracker, the association thresholds were set to 0 for \ac{giou} and 0.9 for embedding similarity. Regarding track management, the minimum number of consecutive associations required to initialize a track ($min\_hits$) was set to 5, while the maximum number of consecutive frames a track could remain unassociated before termination ($max\_age$) was set to 10.

\textbf{Datasets:} The experiments are conducted on the KITTI dataset~\cite{Geiger2012kitti}, a widely adopted benchmark comprising 21 training sequences and 29 test sequences for evaluating multi-object tracking systems. While KITTI is traditionally an automotive dataset, we utilize its pedestrian sequences as a proxy for a mobile robot navigating highly dynamic outdoor environments. Since ground-truth annotations for the test set are not publicly available, the data split proposed in~\cite{weng2020ab3dmot} was adopted to construct training and validation sets from the original training set. Furthermore, the analysis was restricted to the \textit{Pedestrian} category to align with the current focus on human-robot interaction.

\textbf{Evaluation Metrics:} To assess performance in a comprehensive manner, two sets of metrics were utilized:
\begin{itemize} 
    \item \textit{\ac{reid} Metrics:} The \ac{map} was used to evaluate the discriminative power of the appearance models. 
    \item \textit{Tracking Metrics:} To evaluate the tracking system as a whole, \ac{hota}~\cite{luiten2021hota} was selected as the primary metric. Since this study focuses on the impact of \ac{reid}, the Association Accuracy (AssA) sub-metric is also reported to isolate association performance from detection quality (DetA). Additional classical metrics are also included, such as \ac{mota}~\cite{bernardin2008evaluating} and IDF1~\cite{ristani2016performance}, to facilitate comparison with earlier works, and the number of Identity Switches (IDSW) to evaluate track consistency.
\end{itemize}
The system's inference speed, in \ac{fps}, is included to better evaluate the trade-off between accuracy and latency. 

\subsection{ReID Component Analysis}
\begin{figure}[t]
    \centering
    \includegraphics[width=1\linewidth]{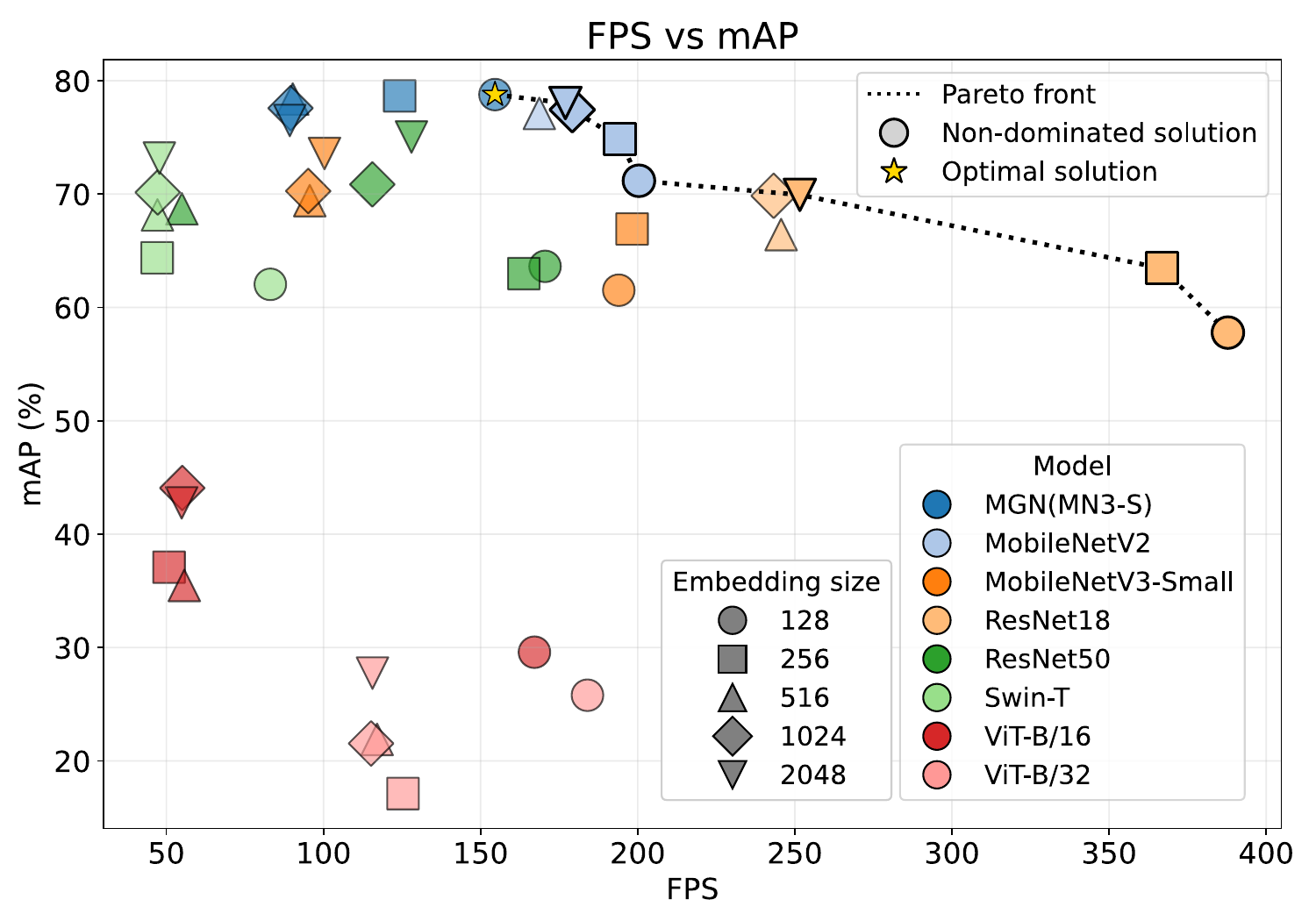}
    \caption{\textbf{Inference Speed (\ac{fps}) vs. Accuracy (\ac{map}) on the Market-1501 dataset.} The marker color indicates the model architecture, while its shape represents the embedding dimension. The \textit{Pareto front} (top-right boundary) highlights the most balanced models.}
    \label{fig:fps_map}
\end{figure}
\subsubsection{Architecture Efficiency Trade-off}
This section will analyze the trade-off between feature representativeness and inference speed for the \ac{reid} module, to determine the optimal backbone for online tracking within the tested architectures. Fig.~\ref{fig:fps_map} illustrates the analysis results. An observable Pareto front is formed mainly by the MGN, MobileNetV2, and ResNet18 architectures, representing the solutions that provide the best trade-off between speed and accuracy. The MGN (using a MobileNetV3-Small backbone) with an embedding size of 128 achieves the highest accuracy, reaching a \ac{map} of approximately 79\% on the Market-1501 dataset. Although it has a complex structure, the inference speed still reaches above 150 \ac{fps}. The MobileNetV2 also achieved adequate results, with mAP scores between 71\% and 78\% and \ac{fps} between 170 and 200, being the most consistent in the group. The ResNet18, while not the most accurate, with \ac{map} between 57\% and 70\%, is the fastest, achieving a remarkable 367 \ac{fps} of inference speed. Vision Transformers (ViT-B/16 and ViT-B/32) displayed weaker performance, achieving lower \ac{map} scores when compared to \acp{cnn}. Notably, the Swin-T transformer architecture did not behave in the same manner, achieving comparable \ac{map} results to \acp{cnn}. This can likely be explained by the limited inductive biases and higher data requirements of ViT-B architectures~\cite{dosovitskiy2020vit}, which are more effectively addressed by Swin-T~\cite{liu2021swin}. Based on this analysis, three configurations were selected to integrate into the 3D tracker: \textit{MGN (128-dim)} as the high-accuracy option, \textit{ResNet-18 (128-dim)} as the high-speed option, and \textit{MobileNetV2 (2048-dim)} as the balanced option.

\begin{figure}[t]
    \centering
    \includegraphics[width=1\linewidth]{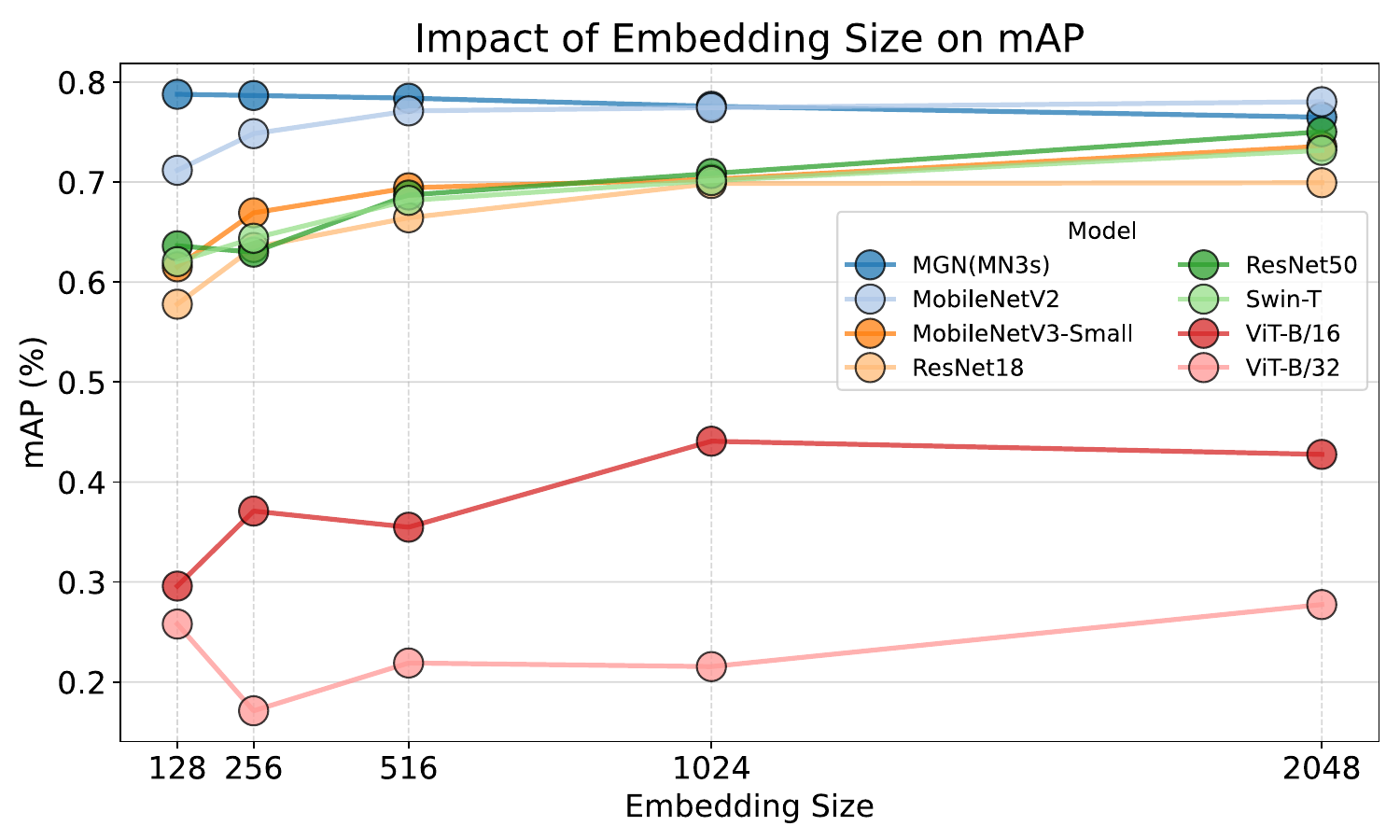}
    \caption{\textbf{Impact of Embedding Size on Discriminability (mAP).} Performance of the selected feature-extracting backbones was evaluated across increasing embedding dimensions $d \in \{128, 256, 512, 1024, 2048\}$.}
    \label{fig:map_embedding}
\end{figure}

Beyond the previous architecture selection, the embedding dimension’s $d$ relation with feature discriminability is also analyzed, as seen in Fig.~\ref{fig:map_embedding}. Two types of behavior are observed. Standard CNNs, such as the ResNets or MobileNets, generally benefit from higher dimensionality. In contrast, the MGN architecture appears to be robust to dimensionality variation, even going against the norm and decreasing slightly with larger embedding sizes. This finding is significant for online tracking applications since it allows for the selection of this heavier network with a more compact embedding size (128), minimizing storage for memory without sacrificing accuracy.

\subsubsection{Impact of Domain Adaptation}
\begin{figure}
    \centering
    \includegraphics[width=1\linewidth]{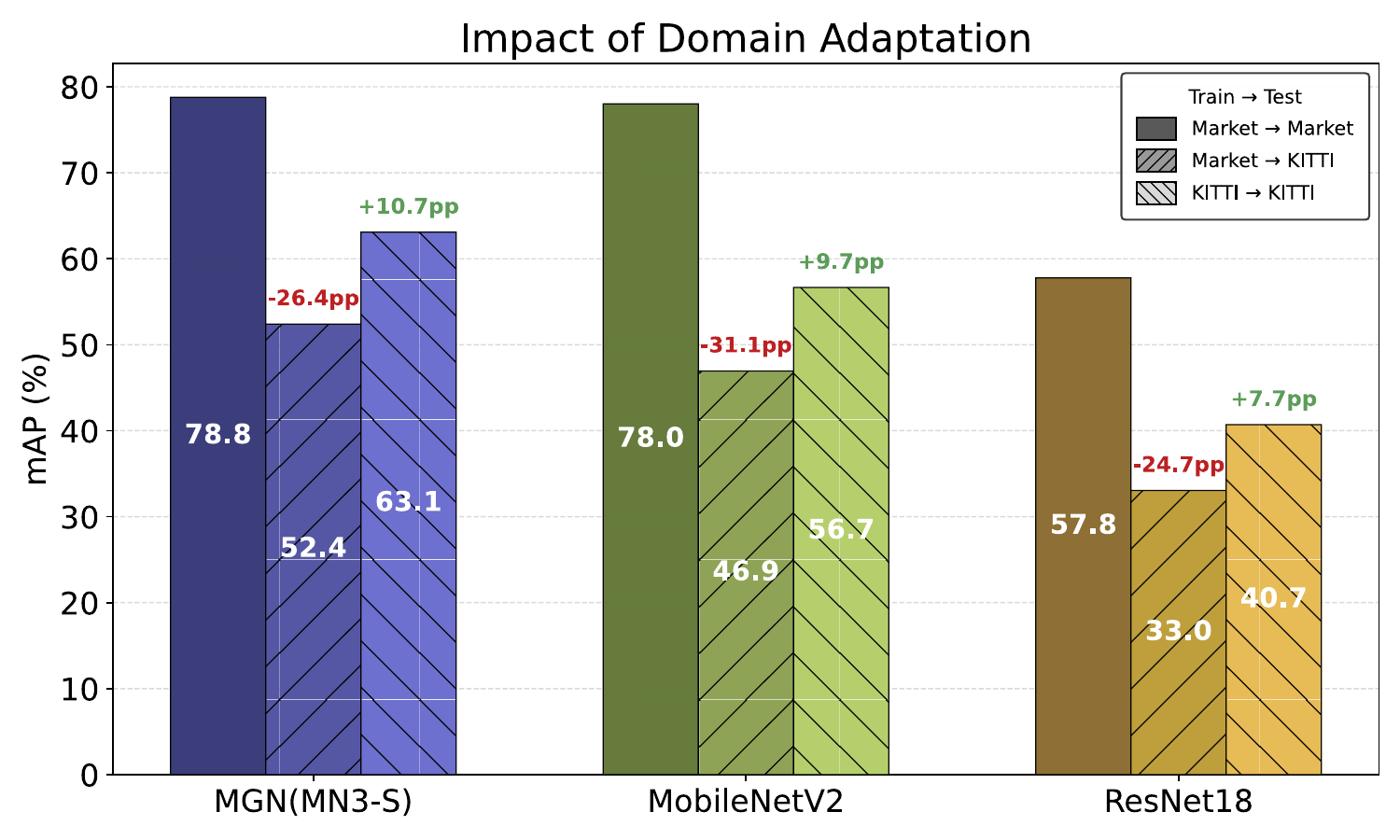}
    \caption{\textbf{Impact of Domain Shift and Fine-Tuning on ReID Accuracy.} The \ac{map} is reported for the three selected architectures: MGN (128-dim), MobileNetV2 (2048-dim), and ResNet-18 (128-dim). The Market$\rightarrow$Market bars represent the baseline performance on the source domain. Direct transfer to the target domain (Market$\rightarrow$KITTI) results in a significant degradation (highlighted in red). Fine-tuning on target domain crops (KITTI$\rightarrow$KITTI) partially recovers this loss (highlighted in green).}
    \label{fig:mAPdomain}
\end{figure}

Fig.~\ref{fig:mAPdomain} highlights the impact of domain adaptation on the three selected models. A direct transfer of models trained on Market-1501 to the KITTI domain results in a significant performance drop, with mAP scores falling by approximately 25-30 percentage points across architectures. This is likely due to the fact that person crops in KITTI are often lower resolution and heavily occluded compared to the clean pedestrian images from Market-1501. However, the implemented fine-tuning strategy effectively shrinks this gap, with MGN recovering 10.7 percentage points.

\subsection{3D Multi-Object Tracking}
In this section, the impact of integrating these ReID models into the full 3D tracking pipeline is thoroughly evaluated.
\subsubsection{Impact of Association Strategy}
Here, the two fusion strategies (Weighted Sum vs. Cascaded) and the appearance-only approach are compared with the geometry-only baseline. Table~\ref{tab:tabass} displays this quantitative evaluation for each of the selected \ac{reid} frameworks, using the Cumulative Moving Average (CMA) as the appearance embedding storage strategy. 

First, relying exclusively on appearance for association (\textit{2d\_emb}) produces clearly inferior results to geometric-only association, with all metrics falling considerably and ID switches increasing more than 10 times over. Second, the naive linear fusion strategy (\textit{w\_sum}) with equal weighing ($w=0.5$) generally degraded the association. While it achieved the lowest number of ID switches for MobileNetV2, with 156, a decrease of 8 from the baseline (164), it produced a noticeable drop in primary tracking metrics such as HOTA, MOTA, and IDF1 across ReID networks. This suggests that this manner of including appearance generated unnecessary noise, dragging down the overall tracking precision in situations where geometric-based matching was sufficient. Finally, the proposed 2-step cascaded strategy (\textit{2-step}) proved to be the most robust approach. It preserved the detection accuracy from the baseline (maintaining MOTA $\approx$ 11\%) while using ReID to recover occluded or disappeared tracks. The MobileNetV2 backbone with 2-step association achieves the highest overall performance (37.67 HOTA, 54.90 IDF1), proving it can offer marginal but consistent gains in tracking consistency (+0.27 IDF1) without introducing the harmful noise observed in weighted sum. However, the addition of ReID comes with a significant latency cost, increasing the total processing time from 57ms to over 100ms per frame.

\begin{table}[t]
\caption{\textbf{Impact of Association Strategy on 3D MOT Performance.} The \textit{Assoc.} column denotes the matching strategy: only geometric (\textit{GIoU}), only appearance (\textit{2d\_emb}), weighted sum (\textit{w\_sum}), or the cascaded matching (\textit{2-step}).}
\label{tab:tabass}
\setlength{\tabcolsep}{2.5pt} % default is 6pt
\renewcommand{\arraystretch}{1.3}
\begin{tabular}{l l c c c c c c}
\noalign{\hrule height 1pt} \hline
Model & Assoc. & HOTA$\uparrow$ & AssA$\uparrow$ & MOTA$\uparrow$ & IDSW$\downarrow$ & IDF1$\uparrow$ & Time$\downarrow$ \\
\hline \hline
-- &  GIoU & 37.58 & 43.43 & \textbf{11.35} & \underline{164} & 54.63 & \textbf{57} \\ \midrule
MGN(MN3s) &  2d\_emb & 22.52 & 17.18 & -9.56 & 1400 & 30.12 & 88 \\
MGN(MN3s) &  w\_sum & 34.86 & 38.51 & 6.40 & 188 & 50.67 & 106 \\
MGN(MN3s) &  2-step & \underline{37.65} & \textbf{43.74} & 10.57 & 166 & 54.75 & 115\\ \midrule
MobileNetV2 &  2d\_emb & 20.06 & 13.90 & -18.50 & 1951 & 27.73 & \underline{85}\\
MobileNetV2 &  w\_sum & 36.40 & 41.59 & 7.91 & \textbf{156} & 53.12 & 102 \\
MobileNetV2 &  2-step & \textbf{37.67} & \underline{43.69} & \underline{11.28} & \underline{164} & \textbf{54.90} & 109 \\ \midrule
ResNet18 &  2d\_emb & 16.81 & 9.90 & -25.08 & 2461 & 22.23 & \underline{85} \\
ResNet18 &  w\_sum & 36.68 & 42.34 & 7.56 & 172 & 54.11 & 104 \\
ResNet18 &  2-step & 37.58 & 43.43 & \textbf{11.35} & \underline{164} & 54.63 & 108 \\ 
\noalign{\hrule height 1pt} \hline
\end{tabular}
\end{table}

\subsubsection{Impact of Feature Memory}
In 3D MOT, object appearance can change rapidly due to occlusions and changes in viewing angle. To mitigate this, two strategies for modeling tracklet appearance were evaluated: a sliding window history (\textit{W}) of size $N$, and an \ac{ema}. Table~\ref{tab:mytable} presents this comparison.

Storing a history of embeddings severely increases computational requirements during the association phase. As $N$ increases from 1 to 50, the total system latency nearly triples (\eg MGN rises from 116ms to 370ms), suggesting that high-$N$ configurations may be unsuitable for real-time applications. Surprisingly, this added cost did not yield performance gains.  For the MGN backbone, the simplest configuration of keeping only the last embedding ($N=1$) actually achieved the highest HOTA (38.24), outperforming all larger history sizes. This suggests that the most recent appearance is often the most reliable, with older embeddings, which were potentially captured from significantly different viewing angles, possibly introducing noise. The EMA strategy might offer an interesting alternative, particularly for the lighter models. For MobileNetV2, EMA achieves the highest HOTA (38.02) for that architecture, and the lowest number of Identity Switches (162) overall, all while maintaining a fast inference time (114ms), which is comparable to the single-frame approach. This indicates that for less discriminative backbones, temporal smoothing might help stabilize the appearance-feature representation. Notably, the ResNet-18 model with the Window strategy ($W$) could not improve over the geometry-only baseline (HOTA 37.58), producing equivalent metrics across all window sizes. This implies that ResNet-18 features may lack the necessary discriminative power to override geometric association.

\begin{table}[t]
\caption{\textbf{Impact of Feature Memory Strategy on Tracking Performance.} Two update mechanisms were compared: fixed-window gallery (\textit{W}) or applying an \ac{ema} with $\alpha=0.9$.}
\label{tab:mytable}
\renewcommand{\arraystretch}{1.3}
\setlength{\tabcolsep}{2pt} % default is 6pt
\begin{tabular}{l c c c c c c c c c}
\noalign{\hrule height 1pt} \hline
Model & \multicolumn{2}{c}{Memory} & HOTA$\uparrow$ & AssA$\uparrow$ & MOTA$\uparrow$ & IDSW$\downarrow$ & IDF1$\uparrow$ & Time$\downarrow$ \\
 & Type & N &  &  &  &  &  &  &  \\
\hline\hline
-- (GIoU) & -- & -- & 37.58 & 43.43 & 11.35 & 164 & 54.63 & \textbf{57} \\ \midrule
MGN(MN3s) & W & 1 & \textbf{38.24} & \textbf{44.95} & 10.93 & \underline{163} & 56.20 & 116 \\
MGN(MN3s) & W & 10 & \underline{38.22} & \underline{44.94} & 10.87 & 166 & \textbf{56.35} & 192 \\
MGN(MN3s) & W & 50 & 38.20 & \underline{44.94} & 10.77 & 166 & \underline{56.34} & 370 \\
MGN(MN3s) & EMA & -- & 37.76 & 44.02 & 10.52 & 167 & 55.31 & 111 \\ \midrule
MobileNetV2 & W & 1 & 37.70 & 43.66 & \underline{11.53} & \underline{163} & 54.86 & \underline{108} \\
MobileNetV2 & W & 10 & 37.79 & 43.90 & 11.47 & \underline{163} & 55.11 & 177 \\
MobileNetV2 & W & 50 & 37.74 & 43.78 & \textbf{11.59} & 164 & 54.99 & 334 \\
MobileNetV2 & EMA & -- & 38.02 & 44.45 & 11.44 & \textbf{162} & 55.75 & 114 \\ \midrule
ResNet18 & W & 1 & 37.58 & 43.43 & 11.35 & 164 & 54.63 & 110 \\
ResNet18 & W & 10 & 37.58 & 43.43 & 11.35 & 164 & 54.63 & 177 \\
ResNet18 & W & 50 & 37.58 & 43.43 & 11.35 & 164 & 54.63 & 361 \\
ResNet18 & EMA & -- & 37.68 & 43.68 & 11.41 & 164 & 54.91 & 112 \\
\noalign{\hrule height 1pt} \hline
\end{tabular}
\end{table}

\section{Conclusion}
In this work, a systematic study was conducted to evaluate the integration of image-based \ac{reid} into online 3D \ac{mot} pipelines. By separating geometric association from appearance modeling, their specific contributions were successfully isolated and quantified.

While geometric heuristics remain highly capable, appearance information serves as an important recovery mechanism for long-term occlusions. Crucially, the integration strategy is a deciding factor. A naive linear fusion of appearance and motion costs degraded performance compared to a geometry-only baseline due to appearance ambiguities. Conversely, a cascaded association strategy—where \ac{reid} is applied only as a secondary stage to recover lost tracklets—delivered the best performance, improving identity consistency without compromising overall precision.

Regarding feature extraction, lightweight \acp{cnn} (such as MobileNetV2) and ReID-specific networks (such as MGN) offered the optimal accuracy-latency trade-off for online applications, outperforming heavier Transformer architectures. Furthermore, long memory structures like sliding windows (with $N>1$) incurred prohibitive computational costs without delivering accuracy gains. Instead, retaining only the latest appearance embedding or applying simple temporal smoothing via \acp{ema} provided the most stable representation.

Ultimately, while RGB-based \ac{reid} enables valuable track-recovery, it introduces a significant latency overhead that is not fully compensated by proportional overall performance gains. Nevertheless, it remains a fundamental addition for critical environments where processing speed can be traded for increased safety and interaction consistency, provided domain-specific fine-tuning and careful association strategies are employed. Future work focusing on models with increased discriminative power and improved domain adaptation will further minimize appearance ambiguity, potentially justifying this computational cost.

\section*{Acknowledgment}
This work has been supported by the Portuguese Foundation for Science and Technology (FCT) through grant ISR-UC UID/00048/2025 (DOI: 10.54499/UID/00048/2025) and by Agenda ``GreenAuto: Green innovation for the Automotive Industry", with reference 02/C05-i01.01/2022.PC644867037-00000013. Eduardo Borges is being supported by the FCT Ph.D. grant 2025.01736.BD.

\balance
\bibliographystyle{IEEEtran}
\bibliography{refs.bib}

\end{document}